\documentclass[sigconf]{acmart}

\usepackage{graphicx}
\usepackage[utf8]{inputenc}
\usepackage[english]{babel}
\usepackage{mathtools}
\usepackage{paralist}
\usepackage{booktabs}
\usepackage{xcolor}
\usepackage{hyperref}
\usepackage{cleveref}
\usepackage{footnote}
\makesavenoteenv{tabular}
\makesavenoteenv{table}

\DeclareMathOperator{\LM}{LM}

\DeclareMathOperator{\train}{train}

\DeclareMathOperator{\val}{val}
\DeclareMathOperator{\test}{test}
\DeclareMathOperator{\degree}{deg}
\DeclareMathOperator{\new}{new}
\DeclareMathOperator{\old}{old}
\DeclareMathOperator{\cls}{cls}
\DeclareMathOperator{\ext}{ext}

\let\cref\Cref
\AtBeginDocument{%
  \providecommand\BibTeX{{%
    \normalfont B\kern-0.5em{\scshape i\kern-0.25em b}\kern-0.8em\TeX}}}

\acmConference[XXX]{XXX}{XXX}{XXX}
\acmBooktitle{XXX}

\begin{document}

\title[LG4AV]{LG4AV: Combining Language Models and Graph Neural Networks for
  Author Verification}

\author{Maximilian Stubbemann}
\orcid{0000-0003-1579-1151}
\affiliation{%
\institution{L3S Research Center and University of Kassel}
\city{Kassel}
\country{Germany}
}
\email{stubbemann@l3s.de}

\author{Gerd Stumme}
\orcid{0000-0002-0570-7908}
\affiliation{%
\institution{University of Kassel and L3S Research Center}
\city{Kassel}
\country{Germany}
}
\email{stumme@cs.uni-kassel.de}

\begin{abstract}
  
  The automatic verification of document authorships is important in
  various settings. Researchers are for example judged and compared by
  the amount and impact of their publications and public figures are
  confronted by their posts on social media platforms. Therefore, it
  is important that authorship information in frequently used web
  services and platforms is correct. The question whether a given
  document is written by a given author is commonly referred to as
  authorship verification (AV). While AV is a widely investigated
  problem in general, only few works consider settings where the
  documents are short and written in a rather uniform style. This
  makes most approaches unpractical for online databases and knowledge
  graphs in the scholarly domain. Here, authorships of scientific
  publications have to be verified, often with just abstracts and
  titles available. To this point, we present our novel approach LG4AV
  which combines language models and graph neural networks for
  authorship verification. By directly feeding the available texts in
  a pre-trained transformer architecture, our model does not need any
  hand-crafted stylometric features that are not meaningful in
  scenarios where the writing style is, at least to some extent,
  standardized. By the incorporation of a graph neural network
  structure, our model can benefit from relations between authors that
  are meaningful with respect to the verification process. For
  example, scientific authors are more likely to write about topics
  that are addressed by their co-authors and twitter users tend to
  post about the same subjects as people they follow. We
  experimentally evaluate our model and study to which extent the
  inclusion of co-authorships enhances verification decisions in
  bibliometric environments.
\end{abstract}




\maketitle

\section{Introduction}
Evaluation of research strongly depends on bibliometric
databases. Today, they are used for the assessment of productivity and
impact of researchers, conferences and affiliations. This implies an
increasing importance of search engines and web services that store,
present and collect bibliometric data. Because of their rising
relevance for the evaluation of the scientific output of individual
authors, it is crucial that the information which is stored and
collected by scholarly search engines, databases and knowledge graphs
is complete and accurate. However, with the rapid growth of
publication output~\cite{bornmann15}, automatic inspections and
corrections of information in bibliometric databases is needed. One of
the major challenges in this area is authorship verification (AV),
which aims to verify if a document is written by a specific author. In
general, AV is widely investigated~\cite{koppel04, halteren04,
  halvani19}. Available approaches range from the use of hand-crafted
features which are based on lexical or syntactic
patterns~\cite{hurlimann15, pothas20, jankowska14} to methods that
make use of neural networks and language models (LMs)~\cite{bagnall15,
  barlas20}.

A majority of existing work handles author verification by capturing
writing styles~\cite{castro15, halvani19}, assuming that they are
unique among different authors. This assumption does not hold in
environments where the available texts are short and contain uniform
language patterns. An example of this is given by verification tasks
for scientific documents. In such settings, the availability of full
texts is rare because bibliometric databases often contain only
abstracts and titles. In such scenarios the variety of writing styles
and linguistic usage is rather limited.  Hence, methods that are
solely based on stylometric features are less promising.

Additionally, the focus in AV research is on documents with one
author, while verification of multi-author documents is seldom
done. In such scenarios, the information about known multi-authorships
can enhance the verification process because it provides a graph
structure between the authors which is meaningful with respect to the
verification process. There are many scenarios where verification
decisions can benefit from such graph structures. For example,
scientific authors are more likely to write papers that would also fit
to their co-authors and twitter users are expected to post about the
same topics as the persons they follow. The incorporation of such
graph structures is rarely investigated.

Most of the current approaches are based on an a setting which breaks
the AV problem down to either ``Are the documents $d_1$ and $d_2$ by
the same author?'' or ``Is the unknown document $d$ from the same
author as the set of known documents $D$?''. Here, $D$ is assumed to
be of small cardinality. In the
PAN@CLEF\footnote{\url{https://pan.webis.de/}} tasks on AV, which most
approaches are focused on, the set of known documents for each unknown
document was never larger then $10$ elements. Hence, the developed
methods are in general not fitted to scenarios where larger sets of
known documents are possible. For example, established researches can
have hundred publications or more which can reflect a variety of
different topics that they have worked on over time. Thus, a large
amount of publications of these research can be relevant for the
verification of potential papers. This makes approaches unfeasible
which need to compare or combine the unknown document explicitly with
the known documents.

Here we step in with LG4AV. Our novel architecture combines language
models and graph neural networks (GNNs) to verify whether a document
belongs to a potential author. This is done without the explicit recap
of the known documents of this author at decision time which can be a
bottleneck with respect to the computation time. This is especially
true for authors with a large amount of known documents. Additionally,
LG4AV does not rely on any hand-crafted stylometric features.

By incorporating a graph neural network structure into our
architecture, we use known relations between potential authors to
enhance the verification process. In this way, we are able to account
for the fact that authors are more likely to turn to topics that are
present in their social neighborhood.
We experimentally evaluate the
ability of our model to make verification decisions in bibliometric
environments and we review the influence of the individual components
on the quality of the verification decisions.
Applications of LG4AV to other data sources where texts are connected
with graph information, such as for example social networks, are
possible. Our contribution is as follows.
\begin{itemize}
\item We study authorship verification in a setting where information
  of known documents is only used at training time. An explicit recap
  of the known documents to verify authorships of
  specific authors with new documents is not needed.
\item We present LG4AV, a novel network architecture that incorporates
  language models and graph neural networks. LG4AV does not depend on
  any stylometric hand-crafted features and allows to incorporate
  meaningful relations between authors into the verification process.
\item We evaluate LG4AV in bibliometric environments
  and study how the different forms of information and the
  different parts of the module enhance the verification process.
\item We additionally investigate to which extent LG4AV is
  capable of verifying potential publications of authors that were not
  seen at training time.
\end{itemize}

LG4AV is available at \url{https://github.com/mstubbemann/LG4AV}.

\section{Related Work}
\label{sec:rel}

Authorship verification is a commonly studied problem. PAN@CLEF
provides regular competitions in this realm. However, their past
author verification
challenges were based in a setting where either small samples of up to
ten known documents for each unknown document were provided
(2013-2015) or pairs of documents were given where the task was to
decide whether they were written by the same person (2020). Both
scenarios are not applicable to our situation, where the amount of
known documents of an author strongly varies and can reach up to
hundreds.

Many well-established methods for author verification develop specific hand-crafted
features that capture stylometric and syntactic patterns of
documents. For example,~\cite{hurlimann15} uses features such as
sentence-lengths, punctuation marks and frequencies of n-grams to make
verification decisions. While the use of n-grams was already studied
in earlier works such as~\cite{kevselj03}, there are still recent
methods that build upon them, such as~\cite{pothas20}. Here, the
authors use them to determine similarities between known, unknown and
external documents to compute verification scores. Another well
established approach is given by~\cite{koppel07} where the authors
successively remove features and observe how this reduces the
distinction between two works. This approach is still known to be the
gold standard~\cite{pothas20, bevendorff19}. Despite its advantages,
it is known to perform worse on short texts. Therefore,
\cite{bevendorff19} proposes a modification that is also applicable to
shorter texts. However, in their work the authors experiment with documents
of 4,000 words per document, which is still much longer then abstracts
of scientific publications.

Recently, methods based on neural network architectures emerged. For
example,~\cite{bagnall15} uses a recurrent neural network with
classifier heads on top for the individual potential authors. Building
up on this work,~\cite{barlas20} proposes to replace the recurrent
network by a pre-trained language model. Note, that both of these
approaches need to train head layers for each individual author. This
makes them unpractical for verification in bibliometric databases
where information of thousands of authors has to be stored.

Author verification has been applied to many scenarios. While earlier
studies often focused on books~\cite{koppel07, kevselj03}, recent
methods consider also online sources such as mails, social media data
and forum postings~\cite{brocardo15, boenninghoff19-2,
  halvani16}. However, author verification for bibliometric data
is rarely done. One of the few works that experiments with
bibliometric data is~\cite{halvani19}, which uses full text of a small
subset of authors and ignores co-authorship relations. Most other
works that deal with bibliometric data tackle the closely related
problem of \emph{authorship attribution} (AA), i.e., with questions of
the kind ``who is the author of $d$'' instead of ``is $a$ author of
$d$''. For example,~\cite{hill03, bradley08} do author verification of
research papers only on citation information while~\cite{caragea19}
also incorporates full texts.

Most of the work that combines
graph information and text features in the realm of bibliometric
data is restricted to the paper level~\cite{cohan20, hamilton17,
kipf17}. Even if there are few works that use text information and
co-author connections, they tackle other problems then AV, such as
link prediction or author clustering~\cite{ganesh16, wang21}.
One of the few works in the realm of AV that explicitly takes into
account that research papers are multi-author documents
is~\cite{sarwar20}. In their work, the authors derive a similarity
based graph structure of text fragments for authorship attribution for
multi-author documents. In contrast, our aim is to incorporate past
co-authorship relations to verify potential authorships.

In the realm of
bibliometric data, many works also have focused on author
disambiguation which has been tackled with the use of graph
information~\cite{kang09, xu18} or text information~\cite{song07}. However, in contrast to AV, author
disambiguation does not tackle the problem of identifying authors of
documents but the related problem of distinguishing authors with
(nearly) equal names.

\section{Combining Language Models and Graph Neural Networks for
  Author Verification}
\label{sec:lg4av}
In the following, we formulate the problem we consider, introduce
our architecture and discuss the individual components of LG4AV.
For the rest of this work, we assume vectors $v \in \mathbb{R}^m$ to be
row-vectors. For a matrix $M \in \mathbb{R}^{m \times n}$, we denote by $M_i$
the $i$-th row of $M$ and with $M_{i,j}$ the $j$-th entry of the $i$-th
row of $M$.

\subsection{Problem}
\label{sec:prob}

Let $t$ be a fixed time point and let $G=(A,E)$ be a graph with
$A=\{a_1, \dots, a_n\}$ being a set of authors and
$E \subset \binom{A}{2}$ a set of undirected and unweighted edges
that represent connections until $t$. Additionally, let
$D$ be a set of documents. Let, for all authors $a \in A$, be
$D(a) \subset D$ the set of their known documents until $t$.  Let $U$
be a set of documents created after $t$ with unknown authorships. The
goal is to verify for a set $P \subset A \times U$ of potential
author-document pairs, whether for each $(a, u) \in P$ $a$ is an
author of the unknown document $u$. More formally, the aim is to find
for each tuple a verification score $f(a, u) \in [0,1]$.

Hence, we aim to infer from the data present at \emph{training time
}information about authors to verify potential documents of them that
occur at \emph{testing time}. Our formulation differs from the usual
setting where the problem is either broken down to sequences of pairs
$(D_i, d_i)_{i=1}^l$ where the task is to determine for each
$i \in\{1,\dots, l\}$ if the \emph{unknown document} $d_i$ is from the
same author as the set of \emph{known documents} $D_i$. These settings
are closely connected in the sense that each author $a$ can be
interpreted as the set of his known documents $D(a)$ at training
time. However, approaches adopted to this setting often assume to have
already pairs of known sets and unknown document ~\cite{hurlimann15}
at training time or they explicitly use the set of known documents for
verification~\cite{pothas20, jankowska14, seidman13}. In contrast, we
train a neural network on incorporating information of known documents
of an author to make it possible to make verification decisions at
testing time without explicitly recapping these documents for each
unknown document. This is especially useful in settings where the
amount of unknown documents can be large for some authors. Here, a
need to explicitly use all these documents for every single
verification of an unknown document can be a potential computational
bottleneck.

\subsection{Combining Language Models and Graph Neural Networks for
  Author Verification}
\label{sec:comb}

\begin{figure}[t]
  \centering\includegraphics[width=\columnwidth]{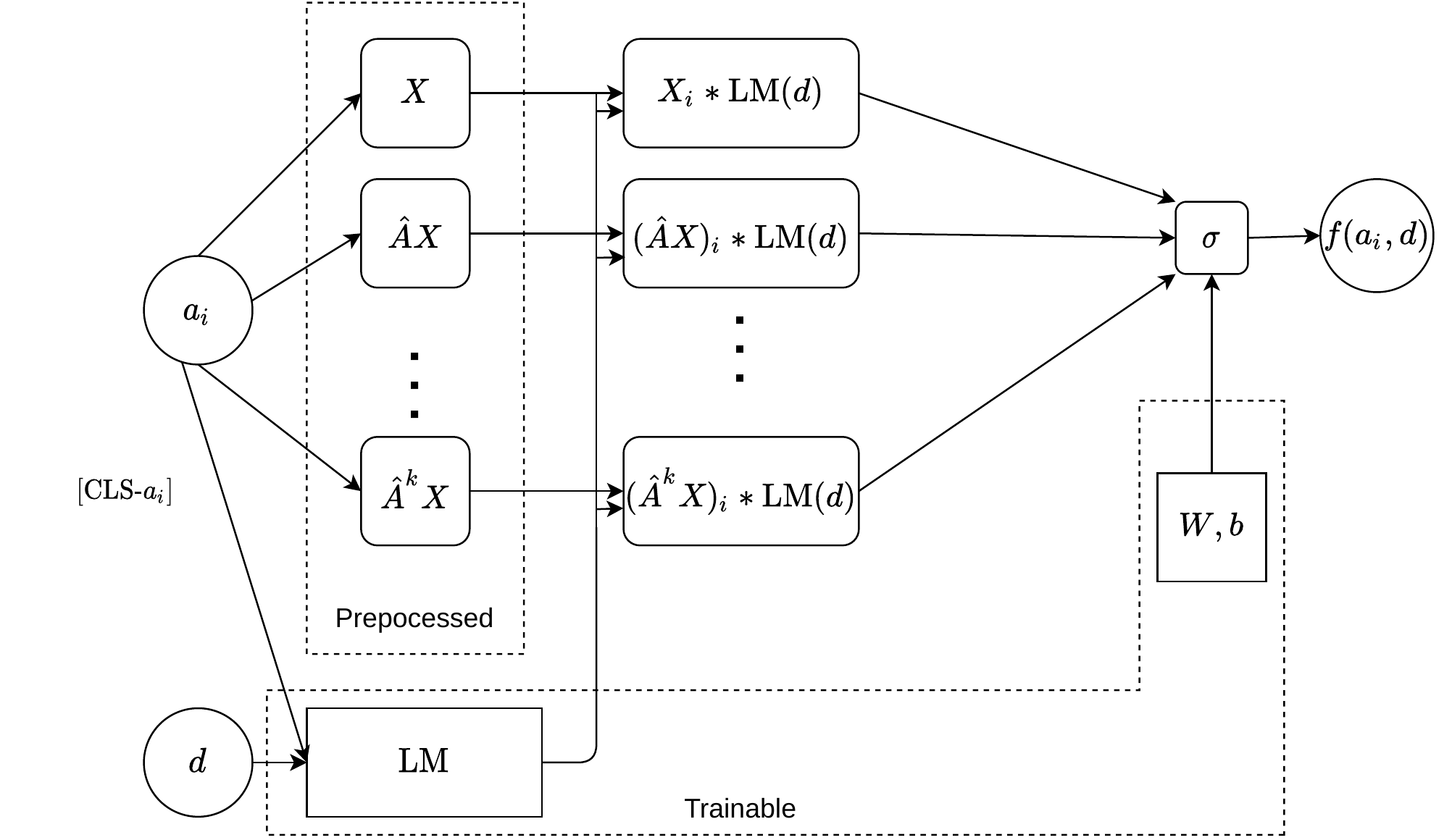}
  \caption{Forward step. LG4AV gets as input an author $a_i$ and a
    document $d$. The author specific cls-token is added to the front
    of $d$ which is then feed through the language model (which is
    BERT in our case). The $i$-th rows of $X,\hat{A}X,\dots,\hat{A}^kX$
    are individually component-wise multiplied with the output of
    $\LM$. The resulting vectors are concatenated and fed through a
    fully connected layer with a sigmoid activation.}
  \label{fig:model}
\end{figure}

We develop an end-to-end model to tackle the author verification
problem. For this, we additionally assume to have for each author
$a \in A$ a vector representation $x_a \in \mathbb{R}^s$. At training
time, our model gets as input pairs $(a,d) \in A \times D$ and is
trained on predicting whether $a$ is author of $D$, i.e.,
$d \in D(a)$. For inference at testing time, the model gets as input
pairs $(a,u) \in P$ and decides whether $a$ is author of $u$ by
computing a verification score $f(a,u)$. For both training and
testing, author-document pairs $(a,d)$ are forwarded through the
network in the following manner. We add a special token to represent
the current author to the beginning of $d$. The resulting document is
then guided through a language model. Additionally, we incorporate a
graph neural network structure by
\begin{inparaenum}
\item[1.)] computing  vector representations of $a$ that
  depend on its graph neighbors,
\item[2.)] combining these vector representations with the
  output of a language model
\item[3.)] and forwarding through a fully connected layer
  to get a verification score.
\end{inparaenum}
In the following, we discuss the individual components and give a
detailed explanation of the network inference. A scratch of LG4AV is
given by \cref{fig:model}.

\paragraph{The language model.}
We choose a neural-network based language model build upon the
transformer architecture~\cite{vaswani17}. More specifically, we
choose the standard BERT~\cite{devlin19} model which we assume the
reader to be familiar with. While this model is originally intended
for sentences, it is possible to feed arbitrary sequences of specific
maximum length (512 tokens for regular BERT models) through the
network. Hence, we feed the full text of the document at once through
BERT and extract the output of the first token (a specific
classification token, denoted by [CLS]). This practice is established
and has already been applied to abstracts of scientific
documents~\cite{cohan20} as well as social media data~\cite{zhu19}. To sum
up, our language model can be interpreted as a map
\begin{equation*}
  \LM: D \cup U \to \mathbb{R}^{m}.
\end{equation*}
To combine the output of the language model with the neighborhood
aggregated author vector representation, we need to ensure that the
output of the language model has the same size as the author
features, i.e., $m = s$. Hence,we extend $\LM$ by a linear
layer on top of the BERT model, if needed.

\paragraph{Author tokens.}
To give the language model information about the current author
$a \in A$, we replace the regular [CLS] token by an author-dependent
classification token [CLS-$a$]. Hence, the information of the current
author is encoded into the input of the language model. Roughly
speaking, if a pair $(a,d)$ is fed through the whole network at
training time, the author information is not only incorporated into
the $\LM$ layers via backpropagation. Instead, $\LM$ also sees $a$ in
the forward step. To nourish from the optimization of the [CLS] token
that was done in the pre-training procedure of BERT, we initialize
for each author $a \in A$ the token-embedding of [CLS-$a$] with the
token-embedding of [CLS].

\paragraph{Choice of the graph neural network.}

We use this paragraph to explain how we incorporate the co-author
information and to discuss the benefits and underlying considerations
of our approach.

The Graph Convolutional Network (GCN)~\cite{kipf17} is a base for many
modern graph neural network architectures. It is a 2-layer neural
network with an additional \emph{neighborhood aggregating} step at the
input and hidden layer. This method leads to problems in
batch-processing. Since the feature vector of each node is merged with
feature vectors of adjacent nodes at the input and hidden layers the
corresponding node vectors have to be in the same batch. This problem
is circumvented by~\cite{kipf17} since they only consider problems
where full-batch training is possible, i.e., training on the whole
feature matrix at once. Various approaches have been proposed to solve
this problem. In~\cite{hamilton17} the authors enable mini-batch
training by sampling the neighborhood of each node. Another approach
is given by~\cite{chiang19} where the authors propose to cluster the
graph into smaller subgraphs which build the batches for
training. While these approaches allow to train neural networks with
smaller batches, they are still not sufficient for our needs since
each batch has to continue a reasonable amount of neighbors and
non-neighbors. This makes it challenging to incorporate BERT which
only allows for small batch sizes. Additionally, in our
architecture the question arises how to combine author-vectors with
output of the language model if the author is only part of the batch
for neighborhood aggregation and not as part of a verification example.

To solve this problem we use a GNN
architecture that only aggregates neighborhood information
\textbf{before} weight matrices are multiplied with feature
vectors. Then, the neighborhood-aggregation can be done once in a
preprocessing step over the full feature-matrix.
Such an approach is for example introduced by~\cite{wu19}. Here, the
authors propose a linear model of the form
$X \mapsto\sigma(\hat{A}^kXW)$, with $\hat{A}$ being a normalized
adjacency matrix and $W$ a trainable weight matrix. In contrast,
SIGN~\cite{rossi20} proposes to have multiple input layers with
different neighborhood aggregations of the form
$X \mapsto \hat{A}_kXW$. The outputs of these layers are then
concatenated, fed through an activation function and another
fully-connected layer. Here, $\hat{A_k}$ can for example be a power of
a normalized adjacency matrix or powers of matrices that are based on
triangles in the graph. While the model in~\cite{wu19} is more simple,
it has the disadvantage that it only uses $\hat{A}^kX$ and not
incorporates X itself. For LG4AV, we use a new model that is strongly
inspired by both of the introduced approaches. Our complete
architecture looks as follows.

\paragraph{LG4AV}
The exact network inference is done in the following manner. Let
$X \in \mathbb{R}^{n \times s}$ be the feature matrix of all authors,
i.e. $X_i = X_{a_i}$.  Let $k \in \mathbb{N}$. Let $\tilde{A}$ be the
adjacency matrix of $G$ with added self loops. For each $i \in \{1,\dots,n\}$, let
$\degree(i):=\sum_{j=1}^n\tilde{A}_{i,j}$ be the degree of $a_i$ in G and
$D \in \mathbb{R}^{n \times n}$ be with $D_{i,i}:=\degree(i)$ and
$D_{i,j} = 0$ if $i \neq j$. Let then
$\hat{A}:=D^{-\frac{1}{2}}\tilde{A}D^{-\frac{1}{2}}$ be the normalized
adjacency matrix. For a given pair $(a_i,d)$ of an author and a
document the network inference is done in the following manner. For
all $l \in \{0,\dots,k\}$, we concatenate the vectors
$v_l(a_i,d):=(\hat{A}^lX)_i*\LM(d)$ to $v(a_i,d):=(v_0,\dots,v_k)$.  Here,
$*$ denotes the element-wise product. To derive a verification score from
this concatenated vector, we feed $v(a_i,d)$ into a fully connected
layer with weight matrix $W \in \mathbb{R}^{s(k+1) \times 1}$, a bias
$b \in \mathbb{R}$ and
sigmoid activation. Hence, the full network inference of LG4AV is
given by the equation
\begin{equation*}
  f(a_i,d)\coloneqq \sigma (v(a_i, d))W + b).
\end{equation*}

\paragraph{Network training.}
For training we use all pairs $(a, d) \in A \times D$ with
$d \in D(a)$ as positive examples. For each $a \in A$ we sample
$|D(a)|$ documents $d \in D \setminus D(a)$ to generate negative
examples. We use binary cross entropy as loss function.

\section{Experiments}
\label{sec:exp}

We experimentally evaluate to which extent LG4AV is able to verify
potential authorships in bibliometric environments.

\subsection{Data Source}
\label{sec:data-src}

We use a data set which consists of publication information of
 AI research communities~\cite{koopmann21}. It is based on data from
DBLP\footnote{\url{https://dblp.org/}} and Semantic
Scholar~\cite{ammar18}.  We use this data source because it
contains titles and abstracts from Semantic Scholar which are needed
for LG4AV but are not included in DBLP. However, the relations
between authors and papers are based on DBLP which is, in our
experience, comparably tidy and accurate. This is crucial for us
since we want to prevent wrong authorship information in
the data itself.

The data set consists of two subsets,
\begin{inparaenum}
\item[1.)] the international AI researchers and all their publications,
\item[2.)] the German AI researchers and all their publications.
\end{inparaenum}
We use the data set of the German AI researchers as our first data
set. As a second data set, we extract all authors with publications at
the KDD conference and all their publications from the data set of the
international AI researchers. We refer to the first as the \emph{GAI}
and to the second as the \emph{KDD} data.

\subsection{Data Preparation}
\label{sec:data-prep}
To evaluate to which extent LG4AV is capable of handling the problem
proposed in ~\cref{sec:prob} we need to generate pairs of authors and
potential co-authors with binary labels for training and testing. We
generate this data for the GAI and the KDD data set in the following
manner.
\begin{itemize}
\item We discard all publications without an English
  abstract.
\item For each publication, we use the title and the abstract as input
  for the language model and concatenate them via a new line char to
  generate the text that represents this publication.
\item We build the co-author graph of all authors until 2015. From
  this co-author graph, we discard all author nodes that do not belong
  to the biggest connected component. Let $A$ be the set of
  authors that are nodes in this graph. We use this graph for the
  neighborhood aggregation. We denote the set of publications until
  2015 of these authors with $D_{\train}$.
\item We generate for all authors and each of their publications a
  positive training example. For all authors, we then sample papers
  from $D_{\train}$ which they are not an author of. We sample in such
  a way that we have for each author an equal amount of positive and
  negative examples.
\item We use data from the year $2016$ for validation. More
  specifically, we use for all authors $a \in A$ all
  publications that they have (co-) authored in 2016 as positive
  validation examples. Let $D_{\val}$ be the set of these
  publications. We sample for all authors papers from $D_{\val}$ that
  they are not author of as negative validation examples. Again, we sample in
  such a way that we have for each author an equal amount of positive
  and negative validation examples.
\item We use all publications from 2017 and newer to generate test
  examples. We create the test data analogously to the training
  examples and
  validation examples.
\end{itemize}

\begin{table}[t]
  \caption{Basic statistics of the data sets. We display from left to
    right: 1.)~The number of authors in $A$, 2.)~the number of
    edges in the co-author graph of these authors at training time,
    3.)~the number of training examples, 4.)~the number of validation
    examples, 5.) the number of test examples.}
  \label{tab:data}
  \centering
  \begin{tabular}[h]{l|lllll}
    \toprule
    & \# Authors & \# Edges & \# Train & \# Validation & \# Test \\
    \midrule
    GAI & 1669 & 4315 & 175118 & 14314& 41558 \\
    KDD & 3056 & 9592 & 254096 & 19976 & 61804 \\
    \bottomrule
  \end{tabular}
\end{table}

Basic statistics of the resulting data can be found
in~\cref{tab:data}. Note, that at validation and testing time, we only use the edges available
  at training time for neighborhood aggregation. Results for new
  co-authorships (and authors!) not seen at training time will be
  evaluated in~\cref{sec:new_authors}.
  
We want to point out that we study AV in a balanced settings, i.e., with an equal
amount of positive and negative examples. This is common practice in
the realm of author verification~\cite{pothas20, halvani19,
  halvani16}.

\subsection{Baselines}
\label{sec:baselines}

Since we study author verification in a novel manner where the
examples are author-paper pairs instead of
pairs of known documents and an unknown document, most of
the existing AV approaches are not directly applicable.

There are existing methods that use language models and neural
networks. However, they often deal with the related problem of
authorship attribution~\cite{barlas20, gupta19}, or they consider the
problem whether pairs of documents are written by the same
author~\cite{boeninghoff19}.

For our experiments, we focus on baselines that can be easily
adapted to author document pairs.
For each baseline,
we explain the needed transformation steps to fit it to our
problem. Additionally, we briefly outline the function principles and
the parameter choices.

\paragraph{N-Gram Baseline.} This baseline is strongly inspired by the
baseline script of the AV challenge of PAN@CLEF 2020.
For all authors we generate a ``superdocument'' by
concatenating all their documents that are available for training.

This  baseline measures the cosine similarity between
the known and unknown documents. More specifically, this baselines
works as follows:  For all pairs of authors $a$ and documents $d$ at validation and
testing time, we measure the similarity between the document and the
superdocument of $a$ to decide whether $a$ is author of $d$. If the
similarity is above a given threshold $t$, we classify the pair as a
positive example. To measure similarities, we build a character-based
$n$-gram TF-IDF vectorizer upon all papers available at training
time. Here, we only use the 3000 features that correspond to the most
frequent $n$-grams across the documents that the vectorizer is build
on. We use this vectorizer to compute for each pair $(a,d)$ at
validation or testing time a vector representation of the superdocument
of $a$ and a vector representation of $d$ and then compute cosine-similarity.

 We use the validation data
to tune the $n$ parameter and the threshold $t$. We tune $n$ via
grid-search on $\{1\dots10\}$ and choose the value that corresponds to
the highest AUC on the validation set. We tune the threshold on the set
$\{\frac{1}{999}i-\frac{1}{999}~|~i \in \{1,\dots,1000\}\}$. Note, that
this means to sample 1000 evenly spaced samples in $[0,1]$. We also
test the median of the distances of the validation examples as
threshold. Since the AUC is independent of this threshold, we tune on
the validation F1 score after the best $n$ is chosen.

\paragraph{GLAD~\cite{hurlimann15}.} This method is intended for pairs
of the form $(D,d)$ where $D$ is a set of documents and $d$ is a
single document. The question is whether $d$ is of the same author
as $D$. Note, that GLAD needs such pairs already for training. Since
our data consists of pairs $(a,d)$ with $a$ an author and $d$ a
document, we build training examples for GLAD in the following
manner. Let $P_{\train}$ be the set of all author-document pairs available
at training time and let, for all $(a,d)$, be $l_{(a,d)} \in \{0, 1\}$
the label of that pair. For each author $a$ we collect the set
$D_{a,+}\coloneqq\{d~|~ (a,d) \in P_{\train},
l_{(a,d)}=1\}=\{d_{a,+,0},\dots,d_{a,+,m}\}$ of positive and the set
$D_{a,-}\coloneqq\{d~|~ (a,d) \in P_{\train},
l_{(a,d)}=0\}=\{d_{a,-,0},\dots,d_{a,-,m}\}$ of negative training
examples. For each $i \in \{0,\dots,m\}$, we feed GLAD with the
training examples $(D_{a,+}\setminus\{d_{a,+,i}\},d_{a,+,i})$ with a
positive label and $(D_{a,+}\setminus\{d_{a,+,i}\},d_{a,-,i})$ with a
negative label.
For validation and testing, we replace pairs $(a,d)$ by
 $(D_{a,+},d)$.

GLAD works as follows. For each pair $(D,d)$ a vector representation
is computed that consists of features that are solely build on $D$ or
$d$, such as for example average sentence length and joint features
which are build on $D$ and $d$, such as entropy of concatenations of
documents of $D$ and $d$. This vector representations are then fed
into a support-vector machine. While the authors of~\cite{hurlimann15}
uses a linear support-vector machine with default parameter setting of
scikit-learn~\cite{pedregosa11}, we enhance GLAD by tuning the
$c$-parameter of the support-vector machine and additionally
experiment with radial kernels where we tune the
$\gamma$-parameter. For both parameters, we grid search over
$\{10^{-3},\dots, 10^{3}\}$. Again, we report the test results for the
model with the best AUC on the validation data.

\paragraph{RBI~\cite{pothas20}.}
The ranking-based impostors method decides for a pair $(D, d)$ if they
are from the same author with the help of a set $D_e$ of external
documents. To use exactly the information available for training
and thus have a fair comparison, we use $D_{a,+}$ as the known
documents and $D_{a,-}$ as the external documents for each pair
$(a,d)$.

By studying for each $d_i \in D$ how many documents of $D_e$ are
closer to $d$, the impostors method computes a verification score
where pairs with higher scores are more likely to be positive
examples.  To compute vector representations for documents, we stick
to the procedure in~\cite{pothas20}. We choose the following
parameters for RBI. We grid search $k \in \{100, 200, 300, 400\}$,
choose cosine similarity as the similarity function and select the
aggregation function between mean, minimum and maximum function. For
the meaning of the parameters, we refer to~\cite{pothas20}. We use the
AUC score on the validation data to choose the best parameters. To
derive binary predictions from the verification scores, we use the
median of all verification scores from the validation data. In
contrast to GLAD, RBI does not need any pairs $(D,d)$ for a training
procedure. Hence, verification scores can be directly computed for
validation and testing data.

\subsection{Model Details}
\label{sec:params}

We implement LG4AV in PyTorch with Lightning and use the Hugging Face Transformers
library for the language model.  For our model, we stick to standard
parameters in the context of GNNs and BERT. We train LG4AV
for 3 epochs.  After the element-wise multiplication of the BERT
output and the text features, we dropout with probability of $0.1$. We
use the ADAM optimizer with weight decay of $0.01$ and a learning rate
of $2*10^{-5}$ with linear decay. We do not use warm-up steps as it
hurts performance. We use $4$ as batch size and do gradient
accumulation of $4$ for an effective batch size of $16$. We set $k=2$
which is a common choice in the realm of GNNs. As BERT
fine-tuning is known to be unstable~\cite{mosbach21, zhang21}, we do
$10$ runs for LG4AV and report mean scores.

To generate
text features for each author, we feed all their documents through the
not fine-tuned BERT model and build the mean point vector of the vector
representations of the [CLS]
tokens. At earlier stages, we also experimented with
LSA~\cite{deerwester90} and doc2vec~\cite{le14} features and witnessed
only negligible differences in performance.

In bibliometric settings, SciBert~\cite{beltagy19} would be the
natural choice for the BERT model of LG4AV as it is trained on
scientific text. However, since SciBert is trained on the Semantic
Scholar Corpus which is also incorporated in the data sets used for
our experiment, it can not be ruled out that SciBert is trained on
texts included in our test data. Hence, to have a fair comparison to
the baselines, we use the ``regular'' BERT base uncased model.

\subsection{Results and Discussion}
\label{sec:results}
\begin{table}[t]
  \caption{Results. We report for all models the AUC-Score, the
    accuracy and the F1-score.}
  \label{tab:results}
  \centering
  \begin{tabular}[h]{l|l|l|l|l|l|l}
    \toprule
    &\multicolumn{3}{l|}{GAI} & \multicolumn{3}{l}{KDD} \\
    & AUC & ACC &F1 &AUC & ACC &F1 \\
    \midrule
    N-Gram &.8624&.7874& .7817&.7592&.6887& .7008\\
    GLAD &.8328&.7500&.7150&.7322&.6698&.6200\\
    RBI & .8251& .7478&.7391&.7452&.6823&.6647\\
    \textbf{LG4AV} & \textbf{.9247}&\textbf{.8520}& \textbf{.8501} &\textbf{.8522}&\textbf{.7683}&\textbf{.7681}\\
    \bottomrule
  \end{tabular}
\end{table}

\paragraph{Results.}
The results can be found in~\cref{tab:results}.  For both data sets,
LG4AV outperforms all
baselines.  Considering
the baselines, N-Gram leads to the best results. It stands out that the scores for the KDD
data sets are generally lower.

\paragraph{Discussion.}
Our results indicate that LG4AV is comparably well suited for the task
of verifying author-document pairs. Note, that for all authors their
papers until 2015 were used to both build the positive training
examples and for computing their feature vector. Hence, considering
the positive examples, the network is trained on verifying
author-document pairs where the document is additionally used to build
the features of this author. Thus, it is reasonable to expect that the
availability to generalize to unseen documents is potentially
limited. However, the results on the test set, which contains only
documents which where not used to build the features vectors of the
authors, show that this is not the case.

While conducting our experiments, we noticed that validation scores
are consistently higher then the test scores. We attribute this to our
splitting procedure. The validation data is, in terms of time,
``closer'' to the training data then the test data. As the topics
researchers work on change over time an increasing timely distance to
the training data hinders the verification of new documents.

While the performance of LG4AV surpasses all baselines, we find that
it is comparably unstable which is a well known problem for
transformer models~\cite{mosbach21, zhang21}. For some random seeds,
LG4AV does not converge to a reasonable solution at all and results in
AUC scores are around $0.5$. Hence, we recommend to try different random
seeds for training and not rely on one training run.

Lastly, the large gap between the performances on the two data sets is
remarkable. One possible explanation is the fact that the GAI data is
based on German researches from all domains of AI while the KDD data
is limited to authors whose research interests overlaps with the
topics of the KDD conference. Hence, it stands to reason that the
documents of the KDD data set are more topically related. In
consequence, the worse results on the KDD data support our hypothesis
that author verification is mainly not about identifying writing style
of unique authors but about identifying the relevant topics of
authors by capturing important words and formulations.

\subsection{Evaluating the Individual Components of LG4AV}
\label{abl}
\begin{table}[t]
  \caption{Results. We report for all models the AUC-Score, the
    accuracy and the F1-score. We compare the regular LG4AV with
    $k=2$~(LG4AV-2) with a model with $k=0$~(LG4AV-0) and a model with
    $k=2$ with freezed BERT layers~(LG4AV-F).}
  \label{tab:abl}
  \centering
  \begin{tabular}[h]{l|l|l|l|l|l|l}
    \toprule
    &\multicolumn{3}{l|}{GAI} & \multicolumn{3}{l}{KDD} \\
    & AUC & ACC &F1 &AUC & ACC &F1 \\
    \midrule
    LG4AV-F &.8384& .7588& .7747&.7622& .6865& .7101\\
    LG4AV-0 & .9207& .8482& .8477&.8465&.7636&.7645\\
    LG4AV-2 & \textbf{.9247}\footnote{\label{fn:p}
  Significantly outperforms LG4AV-$0$
  with $p=0.01$.} &\textbf{.8520}& \textbf{.8501}
          &\textbf{.8522}\textsuperscript{\ref{fn:p}} &\textbf{.7683}\textsuperscript{\ref{fn:p}}&\textbf{.7681}\textsuperscript{\ref{fn:p}}\\
    \bottomrule
  \end{tabular}
\end{table}

The novelty of LG4AV  lays in the connection of two components, namely
a GNN and a fine-tunable language model, to perform author
verification tasks. Hence, it is crucial to evaluate the individual
components with respect to their influence on the verification
decisions. In order to do so, we run the experiments in \cref{sec:exp}
with two additional LG4AV models.

\paragraph{LG4AV-F} This model coincides with the model used in
\cref{sec:exp} with the difference, that we freeze all BERT parameters
and just train the weight matrix $W$. This allows us to understand if
the process of fine-tuning the BERT parameters is indeed necessary for
successful author verification.

\paragraph{LG4AV-$0$.} Here, the parameter $k$ is set to $0$. Hence,
this model does not use any graph information. It just uses the
individual author features and does not include any neighborhood
aggregation. We use this model to evaluate to
which extent the graph information enhances the verification process.

\paragraph{LG4AV-$2$} The LG4AV model with $k=2$ which was
also used in~\cref{tab:results}. This is our ``regular'' LG4AV
which uses both BERT fine-tuning and neighborhood aggregation.

\paragraph{Procedure}
\label{sec:prod}
We also run $10$ rounds of LG4AV-$0$ and LG4AV-F and report mean
scores. For this runs, we use the same 10 different random seeds for
weight initialization and for shuffling the training data which were
used for LG4AV-$2$. We decided for this approach as early experiments indicated
that the performances of LG4AV-$2$ and LG4AV-$0$ were better for the
same random seeds (and therefore same shuffles of training data). This
means, that the seeds which lead to the higher/lower values for
LG4AV-$2$ generally also lead to better results for LG4AV-$0$.

\paragraph{Results and discussion.}
 The results can be found in \cref{tab:abl}.
Since the results of LG4AV-$2$ and LG4AV-$0$ are very close, we use
statistical significance tests. Based on the procedure explained
in~\cref{sec:prod} and the observation that both models have the
tendency of having better/worse results for the same random seeds, we
decide to use a paired t-test over the 10 runs for significance testing.
 On the KDD data set,
LG4AV-$2$ outperforms LG4AV-$0$ with a significance level of $0.01$ on
all metrics. On the GAI data set, LG4AV-$2$ outperforms LG4AV-$0$ with
a significance level of $0.01$ with respect to the AUC score. Considering accuracy,
LG4AV-$2$ outperform LG4AV-$0$ on the significance level
$p=0.06$. However, for the common value $p=0.05$, it does not. There
is no significant difference for the F1-score with respect to a
reasonable p-value.

To sum up, the results indicate that the incorporation of co-author
information can lead to additional enhancements. Still, the
comparable results
for $k=0$ show that our idea of combining text features with a
language model even works without the addition of co-author
information. On the other hand, if the BERT layers are frozen, the
performance declines considerably. Hence, the fine-tuning of the
language model is an integral point for the successful author
verification with LG4AV.

Comparing the performance of LG4AV-F with the results in~\cref{tab:results},
the frozen models perform on line with GLAD and RBI, which shows
that LG4AV-F is (to some extent) capable of successful author
verification. We especially point that out because the freezing of the
BERT layers significantly decreases the runtime since only the $1+k*768$
parameters of the top layer have to be trained. On a NVIDIA RTX
2060 SUPER, each epoch of LG4AV-F and LG4AV-2 on the GAI data last for
about 40 minutes and for about 2 hour and 15 minutes,
respectively. For the KDD data, one epoch needs about 1 hour and about
4 hours and 15 minutes, respectively. With limited resources
available it would be a reasonable compromise to, for example, train
the whole model for just one epoch and then freeze the layers of the
language model.

\subsection{Author Verification for New Authors}
\label{sec:new_authors}

Until now, we focused on verification cases for authors that were
already seen at training time. However, as LG4AV takes as input examples
an author represented by a feature-vector and a document text to decide whether this document fits to
the feature-vector, it is possible to make verification decisions for
(feature-vectors) of authors which where not seen at training time. We
use this section to evaluate LG4AV on such not seen authors and with
the use of known edges not available at training time.

For this,
we assume to have an additional set of new authors $A_{\new}$ and with
additional known documents $D_{\new}$. Again let for $a \in A_{\new}$
be $D_{\new}(a)$ the set of known documents of this new author and let $U_{\new}$
be a new set of documents of unknown authorship. Additionally we
assume to have a new unweighted,  undirected graph $G_{\new}(A \cup
A_{\new}, E_{\new})$. Hence, we assume to have also edges between the new authors and the old authors. The goal is again, to make
verification decisions
for a set $P_{\new} \subset A_{\new} \times U_{\new}$ .

\paragraph{CLS-token for new authors.}
Note, that for each author $a \in A$ seen at training time, we train a
specific cls token [CLS-$a$] which we place at the beginning of each
document $d$ for each input pair $(a,d)$. Hence, if we want to infer
verification scores for pairs $(a,u) \in P_{\new}$ of new authors and
potential documents, the question arises on how to choose the cls
token [CLS-$a$] as it is not trained. For this, we use the information
of co-author relations to authors that were present at training
time. More specific, we choose the cls token in the following manner.

Let $N_{\old}(a)$ be the set of neighbors of $a \in A_{\new}$
in $G$ that are in $A$. Let for all
$a \in A \cup A_{\new}$ be $\cls(a) \in \mathbb{R}^m$  the vector
representation of [CLS-$a$]. For $a \in A_{\new}$ we set

\begin{equation*}
  \cls(a) \coloneqq
  \begin{cases}
    \frac{1}{|N_{\old}(a)|}\sum_{b \in N_{\old}(a)} \cls(b) &
    \text{if } N_{\old}(a) \neq \emptyset, \\
    \frac{1}{|A|}\sum_{b \in A} \cls(b) & \text{else}.
  \end{cases}
\end{equation*}

To sum up, we set the set the embedding of the cls-token of a new authors
to the mean point vector of the cls-tokens of its neighbors that were
already present at training time. If there are no such old
neighbors, we use the mean point vector of the embeddings of the cls-tokens of all
old authors instead.

\paragraph{Data.}
\begin{table}[t]
  \caption{Basic statistics of the data sets that include unseen authors. We display from left to
    right: 1.)~The number of new authors in $A_{\new}$, 2.)~the number
    of new authors with edges to old authors $a \in A$
    3.)~the average amount of edges to old authors of all new
    authors with at least one edge to old authors, 4.) the number of test examples.}
  \label{tab:data_unseen}
  \centering
  \begin{tabular}[h]{l|lllll}
    \toprule
    & \# Authors & \# Edges Back & $\oslash$ Edges & \# Test \\
    \midrule
    GAI & 35 & 22 & 1.7727 & 514\\
    KDD & 227 & 94 & 2.1383 & 2030 \\
    \bottomrule
  \end{tabular}
\end{table}

For both the GAI and the KDD data set, we choose the set of new
authors $A_{\new}$ in the following manner. For the GAI data set, we choose all
authors, that have their first publication in our former validation
time windows, i.e., in the year 2016. For the KDD data set, we choose all authors,
that have a publication at the KDD conference and which have their
first publication in 2016. For $E_{\new}$, we choose all co-author
edges between authors of $A \cup A_{\new}$ until 2016. Note, that this
also includes new co-authorships between old authors $a,b \in A$ that
first arise in 2016.

For $a \in A_{\new}$ we choose the authored publications of 2016 of
this author as the known documents $D_{\new}(a)$. Let
$D_{\new}\coloneqq \bigcup_{a \in A_{\new}}  D_{\new}(a)$ be the set of the
documents of all new authors in 2016. For the RBI baseline, we sample
for each $a \in A_{\new}$ a set $D_{\new_{\ext}}(a)$ of external documents $a$
has not authored from $D_{\new}$ such that
$|D_{\new}(a)|=|D_{\new_{\ext}}(a)|$. The positive test examples
$D_{\new_{\test}}$ are given by pairs $(a,d)$ where $a \in A_{\new}$
and $d$ is a document authored by $a$ after 2016. Let $D_{\new_{\test}}$
be the set of all documents of all authors $a \in A_{\new}$ after 2016, i.e.,
$D_{\new_{\test}}\coloneqq  \{d~|~\exists a \in A_{\new}: (a,d) \in
D_{\new_{\test}}\}$. For each author $a \in A_{\new}$ we sample
documents that $a$ has not authored from $D_{\new_{\test}}$. For each
positive example for $a$ we sample one negative example. Statistics of
the resulting data can be found in \cref{tab:data_unseen}.

\paragraph{Procedure.}
For all baselines we use the parameters and models for which we
reported scores in~\cref{tab:results} to infer the verification scores
for the new test examples. For LG4AV, we report mean results over the
10 models used in~\cref{tab:results} and~\cref{tab:abl}.

\begin{table}[t]
  \caption{Results for author verification of unseen authors. We
    report for all models the AUC-Score, the accuracy and the
    F1-score.}
  \label{tab:results_unseen}
  \centering
  \begin{tabular}[h]{l|l|l|l|l|l|l}
    \toprule
    &\multicolumn{3}{l|}{GAI} & \multicolumn{3}{l}{KDD} \\
    & AUC & ACC &F1 &AUC & ACC &F1 \\
    \midrule
    N-Gram &.8355&.7276& .6296&.7125&.6276& .4341\\
    GLAD &.8619&.7996&.7785&\textbf{.7421}&\textbf{.6749}&.6643\\
    RBI & .7901& .7257&.7283&.6732&.6172&.6197\\
    \textbf{LG4AV} & \textbf{.8898}&\textbf{.8354}& \textbf{.8465}&.7375&.6671&\textbf{.7045}\\
    \bottomrule
  \end{tabular}
\end{table}

\paragraph{Results and Discussion.}

The results of this experiment can be found
in~\cref{tab:results_unseen}. On the GAI data set, LG4AV outperforms
all baselines. On the KDD data set, GLAD leads to the best
performance with respect to the AUC and accuracy score, followed by
LG4AV. However, LG4AV still outperforms all baselines with respect to
the F1-score. It stands out, that all methods except GLAD perform
remarkably worse for unseen authors
then in the experiment with the same author set for training and
testing. This effect is remarkable strong for LG4AV on the KDD data set.

Having a look at~\cref{tab:data}, it stands out, that in the KDD data
set a smaller ratio of new authors have connections to old
authors ($41 \%$) then for the GAI data ($63 \%$). Hence, it may be
promising to study with more data sets if a general connection between the
ratio of new authors with connection to old authors and the
performance exists. If this is indeed the case, modifications of the generation of the
author-dependent cls-token for new authors would be a potential
starting point for suiting LG4AV especially to verification problems
of authors not seen at training time. 

\section{Conclusion and Outlook}
\label{sec:conclusion}

In this work, we presented LG4AV, a novel architecture for author
verification. By combining a language model with a graph neural
network, our model does not depend on any handcrafted features and is
able to incorporate relations between authors into the verification
process. LG4AV surpasses methods that use
handcrafted stylometric and n-gram text features when it comes to
verification of short and, to some extent, standardized texts, as for
example titles and abstracts of research papers. Hence, LG4AV is
especially helpful to correct authorship information collected
and stored by search engines and online data bases in the scholarly domain.

Future work could include applications of LG4AV to different settings,
as for example authorship verification of tweets.

While LG4AV is generally designed for the verification of potential
authorships of authors seen at training time, it also lead to
comparable good results for verification cases on authors that were
not seen at training time. However, as noted and discussed
in~\cref{sec:new_authors}, the performance drops in such scenarios and
modifications of LG4AV to such tasks would be a useful direction
for future research. Here, modifications of the incorporation of
author information via the author-dependent cls-token could be
promising.

From our observation that LG4AV and the baselines generally perform better
on the validation then the testing data, we concluded that bigger time
spans between training and inference data lead to worse results since
the topics author consider continuously change over time. Building up
on this, it would be interesting to study the influence of the
discarding of old training examples. Does the performance increase
if only recently published papers are considered to verify new potential
authorships?

\begin{acks}
  This work is partially funded by the German Federal Ministry of
  Education and Research (BMBF) in its program ``Quantitative
  Wissenschaftsforschung'' as part of the REGIO project under grant
  01PU17012A. The authors would like to thank Dominik Dürrschnabel and
  Lena Stubbemann for fruitful discussions and comments on the
  manuscript.
\end{acks}

\bibliographystyle{ACM-Reference-Format}
\bibliography{literature}

\end{document}